\documentclass[conference]{IEEEtran}
\IEEEoverridecommandlockouts
\usepackage{cite}
\usepackage{amsthm}
\theoremstyle{definition}

\usepackage{amsmath,amssymb,amsfonts}
\usepackage{threeparttable}
\usepackage{graphicx}
\usepackage{textcomp}
\usepackage{algorithm}
\usepackage{algpseudocode}
\usepackage{enumitem}   
\usepackage{subfigure}
\usepackage{multirow}
\usepackage[dvipsnames]{xcolor}

\definecolor{darkyellow}{rgb}{0.85, 0.65, 0.0}
\definecolor{paleorange}{rgb}{0.95, 0.75, 0.5} % Slightly darker pale orange
\definecolor{palegreen}{rgb}{0.5, 0.85, 0.5}   % Slightly darker pale green

\def\BibTeX{{\rm B\kern-.05em{\sc i\kern-.025em b}\kern-.08em
    T\kern-.1667em\lower.7ex\hbox{E}\kern-.125emX}}

\usepackage{hyperref}
    
\begin{document}

\title{FAME: Introducing Fuzzy Additive Models for Explainable AI \\
\thanks{This work was supported by MathWorks\textsuperscript{\textregistered} in part by a Research Grant awarded to T. Kumbasar. Any opinions, findings, conclusions, or recommendations expressed in this paper are those of the authors and do not necessarily reflect the views of MathWorks, Inc.}
%}
}

\author{\IEEEauthorblockN{Ömer Bahadır Gökmen} 
\IEEEauthorblockA{\textit{AI and Intelligent Systems Lab.} \\
\textit{Istanbul Technical University}\\
%\textit{AI2S Lab.}\\
Istanbul, Türkiye \\
{gokmeno19}@itu.edu.tr}\\
\and
\IEEEauthorblockN{Yusuf Güven} 
\IEEEauthorblockA{\textit{AI and Intelligent Systems Lab.} \\
\textit{Istanbul Technical University}\\
%\textit{AI2S Lab.}\\
Istanbul, Türkiye \\
guveny18@itu.edu.tr}\\
\and
\IEEEauthorblockN{ Tufan Kumbasar} 
\IEEEauthorblockA{\textit{AI and Intelligent Systems Lab.} \\
\textit{Istanbul Technical University}\\
Istanbul, Türkiye \\
kumbasart@itu.edu.tr}\\
}

\maketitle

\begin{abstract}
In this study, we introduce the Fuzzy Additive Model (FAM) and FAM with Explainability (FAME) as a solution for Explainable Artificial Intelligence (XAI). The family consists of three layers: (1) a Projection Layer that compresses the input space, (2) a Fuzzy Layer built upon Single Input-Single Output Fuzzy Logic Systems (SFLS), where SFLS functions as subnetworks within an additive index model, and (3) an Aggregation Layer. This architecture integrates the interpretability of SFLS, which uses human-understandable if-then rules, with the explainability of input-output relationships, leveraging the additive model structure. Furthermore, using SFLS inherently addresses issues such as the curse of dimensionality and rule explosion. To further improve interpretability, we propose a method for sculpting antecedent space within FAM, transforming it into FAME. We show that FAME captures the input-output relationships with fewer active rules, thus improving clarity. To learn the FAM family, we present a deep learning framework. Through the presented comparative results, we demonstrate the promising potential of FAME in reducing model complexity while retaining interpretability, positioning it as a valuable tool for XAI.

\end{abstract}


\section{Introduction}
Deep Learning (DL) has become essential across various applications, yet it faces transparency challenges, especially in mission-critical applications. Often characterized as "black boxes," they offer little insight into their decision-making processes, leading to the need for Explainable AI (XAI) \cite{minh2022explainable}. 

% Although DL excels at identifying complex relationships between input features and outputs in large-scale datasets, its opacity has prompted the development of alternative approaches. One such solution combines additive index models with neural networks (e.g., neural additive models) to explain input-output relations \cite{NEURIPS2021_251bd044, 9149804, gaminet}. 
Neural additive models, an XAI solution, address DL opacity by combining additive index models with neural networks to improve interpretability, while still excelling at capturing complex input-output relationships in large-scale datasets \cite{NEURIPS2021_251bd044, 9149804, gaminet}. These models process each input feature separately through subnetworks, adhering to the principles of the additive index model to improve interpretability \cite{xnn}. However, as underlined in \cite{extract}, there is a need to define clear rules-based structures to translate model output behaviors into human-understandable forms, thereby enhancing interpretability.

% Due to their human-centric linguistic if-then rule-based structure, Fuzzy Logic Systems (FLSs), inherently interpretable by design as widely acknowledged in the fuzzy logic community, are a promising avenue for XAI  \cite{CORDON2011894, 8481251, 8610271, 9430516}.
% Fuzzy Logic Systems (FLSs), inherently interpretable by design through their human-centric linguistic if-then rule-based structure, are a promising tool for XAI, as widely acknowledged in the fuzzy logic community \cite{CORDON2011894, 8481251, 8610271, 9430516}. 

Fuzzy Logic Systems (FLSs), with their human-centric linguistic if-then rule-based structure, are positioned as a highly interpretable and promising solution for XAI, a view widely supported in the fuzzy logic community \cite{CORDON2011894, 8481251, 8610271, 9430516}. Yet,  learning FLSs with high-dimensional data leads to challenges such as the curse of dimensionality and rule explosion \cite{curse, HTSK}. Thus, hybrid approaches combining DL and FLSs are explored to develop XAI \cite{7047917, 7377034, 8015718, 8491679}. However, they often fail to fully leverage the interpretability advantages of FLS \cite{10.1007/978-3-030-04070-3_1}.

\begin{figure}[t]
 \includegraphics[width=0.48\textwidth]{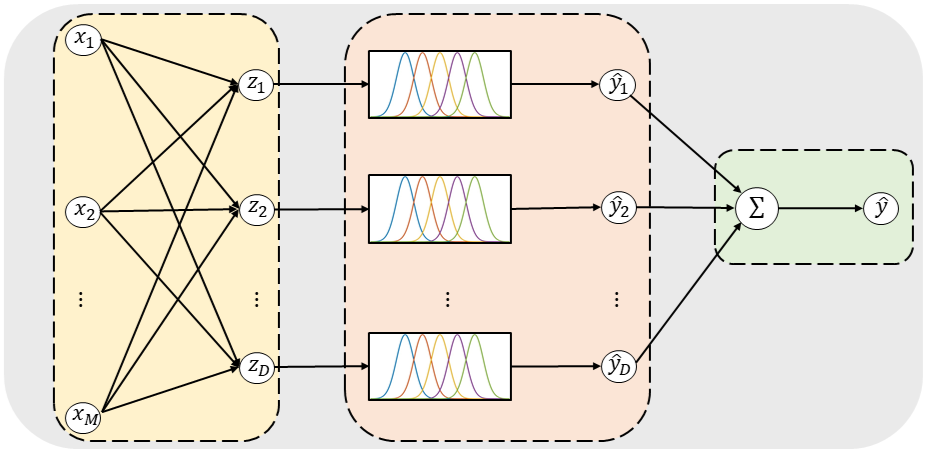}
\caption{FAM/FAME is composed of three main layers: \textit{\textcolor{darkyellow}{Projecton Layer}} maps the original input space $\boldsymbol{X}$ to the feature space $\boldsymbol{Z}$. In the \textit{\textcolor{paleorange}{Fuzzy Layer}}, each reduced dimension \( z_i \in \boldsymbol{Z} \) is passed through an SFLS. The \textit{\textcolor{palegreen}{Aggregation Layer}} combines the outputs of the Fuzzy Layer by summing them.}  
\label{fig: FAM/FAME}
\end{figure}

In this study, we introduce the Fuzzy Additive Model (FAM) and FAM with Explainability (FAME) for XAI. As shown in Fig. \ref{fig: FAM/FAME}, we develop a model with 3 layers: (1) a Projection Layer (PL) for compressing the input space via a linear kernel, (2) a Fuzzy Layer (FL) defined with Single Input-Single Output FLS (SFLS) as subnetworks in additive index models, and (3) an Aggregation Layer. The XAI model combines the interpretability of SFLS, characterized by human-understandable if-then rules, with the explainability of input-output relations through its additive nature. Moreover, by deploying SFLS, the issues of the curse of dimensionality and rule explosion are naturally mitigated, as the design inherently constructs a 1D mapping. To improve the interpretability of the antecedent space, we present a method for sculpting antecedent Membership Functions (MFs) in FAM, i.e. transforming it into FAME. We show that FAME captures the input-output mapping with a minimal active set of rules. We also present a DL framework for learning FAM and FAME.

To evaluate effectiveness, we conduct comparative analyses between FAM/FAME and established Multi Input-Single Output FLS (MFLS). We show that while FAM exhibits superior performance, FAME distinguishes itself through transparency in its antecedent MFs, providing comparable performance. This interpretability is achieved with only a marginal reduction in accuracy, positioning FAME as a solution for XAI.

\section{Fuzzy Additive Models: FAM and FAME}\label{FAM_intro}
This section presents the FAM and FAME depicted in Fig. \ref{fig: FAM/FAME}, detailing their framework and properties.

\subsection{Inference Structure}
FAM/FAME is composed of the following three main layers. 

\subsubsection{Projection Layer}
The PL functions as a linear kernel that maps the input space $\boldsymbol{X} \in \mathbb{R}^{M \times 1}$ to a more representative, low-dimensional space $\boldsymbol{Z} \in \mathbb{R}^{D \times 1}$ as follows: 
\begin{equation}\label{projection_layer}
\boldsymbol{Z} = \boldsymbol{W}\boldsymbol{X} + \boldsymbol{b},
\end{equation}
where $\boldsymbol{W} \in \mathbb{R}^{D \times M}$ is a weight matrix and $\boldsymbol{b} \in \mathbb{R}^{D \times 1}$ is a bias vector. The main aim of the layer is to extract a low-dimensional (i.e. $M \geq D$) and compact representation space. The inherent linearity of the transformation facilitates interpretability by allowing the weight matrix $\boldsymbol{W}$ to serve as a direct representation of feature importance.

% This layer aims to reduce the dimensionality of the input space from $M$ to $D$ (with $M \geq D$). Let $\boldsymbol{X} \in \mathbb{R}^{M \times 1}$ denote the original input vector and $\boldsymbol{Z} \in \mathbb{R}^{D \times 1}$ represent the reduced-dimensional output. Specifically, the transformation is given by:

% The projection layer functions as a linear kernel, mapping the input space to a more representative, lower-dimensional domain. It extracts essential features $\boldsymbol{Z}$ from $\boldsymbol{X}$, offering a compact representation for the subsequent inference processes of FAME.

% The projection layer effectively acting as a linear kernel that maps the input space into a more representative, lower-dimensional domain.

% After this dimensionality reduction, the projection layer extracts essential features from the original input, providing a suitably processed representation for the subsequent inference procedures of FAME.

\subsubsection{Fuzzy Layer}
In this layer, each reduced input dimension $z_i \in \boldsymbol{Z}$ (for $i=1,\ldots,D$) is processed by SFLSs as:
\begin{equation}\label{SFLS}
\hat{y}_i = g_i(z_i; \theta_i),
\end{equation}
\noindent where $g_i(\cdot)$ is the $i^{th}$ SFLS parameterized with $\theta_i$ and $\hat{y}_i$ is its output. The SFLS offers the advantage of being a 1-D mapping, making it easier to interpret than MFLSs, where input interactions can complicate the interpretation. 

\subsubsection{Aggregation Layer}
It combines the outputs of FL $\hat{y}_i$ as: 
\begin{equation}
\hat{y} = \sum_{i=1}^{D}\hat{y}_i.
\end{equation}
where $\hat{y}$ is the model output. This layer is inherently explainable, as it aggregates the outputs of each SFLS, making it easy to trace how each $\hat{y}_i$ contributes to the final prediction. 

% . Each subnetwork provides one scalar output , and these are summed together to produce the final system output :

% By integrating $\hat{y}_i$'s, the aggregation layer reconstructs a comprehensive prediction from $Z$, completing the inference process of the CDR-GAM-FUZZY structure.

\subsection{SFLSs: Inference and Feature partitioning}
% In the CDR-GAM-FUZZY structure, each FLS subnetwork (either T1-FLS or IT2-FLS) operates as a single-input single-output (SISO) system. In this context, the dimension $D$ is no longer explicit within the FLSs, as each subnetwork handles only one input $z_i$. 
This section provides the inference and properties of the SFLSs. For simplicity, we drop the subscript $i$ in the input $z_i$ and output $y_i$ and denote them as $z$ and $y$, respectively.  

\subsubsection{Inference} The rule structure of the SFLSs composed of $P$ rules $(p=1,2,\ldots,P)$ is as follows:
\begin{equation}\label{rule_T1_single}
R_{p}: \text{If } z \text{ is } A_{p} \text{ Then } \hat{y} \text{ is } y_p,
\end{equation}
where the consequent part of the rules are defined with:
\begin{equation}\label{consequent_layer_single}
    y_p = a_{p}z + a_{p,0}.
\end{equation}

The mapping of SFLS $g(\cdot)$ is relatively simpler to its multi-input part since the firing strength of the $p^\text{th}$ rule $f_p(z)$  reduces to the MF grade of $A_{p}$, i.e. $f_p(z)= \mu_{A_{p}}$ \cite{siso_fls}. It is defined as follows: 
\begin{equation}\label{y_T1_single}
g(z; \theta) = \frac{\sum_{p=1}^P \mu_p(z)\,y_p}{\sum_{p=1}^P \mu_p(z)} 
\end{equation}

\subsubsection{Sculpting the antecedent space}\label{Gauss2MF_Intro}
In the literature on FLSs, the Gaussian MF is the most widely used due to its ability to model uncertainty and handle smooth transitions between membership levels \cite{guven}. For this reason, we partition the antecedent space of FAM with Gaussian MFs, defined as: 
% \begin{equation}\label{T1_antecedent_single}
% \mu_{A_{p}}(z) =  \exp\left(-\frac{(z - c_{p})^2}{2 \sigma_{p}^2}\right)
% \end{equation}
\begin{equation}\label{T1_antecedent_single}
           \mu_{A_{p}}(z)=\exp \left(-\left(z-c_{p}\right)^{2} / 2 (\sigma_{p})^{2}\right)
\end{equation}
where $c_{p}$ and $\sigma_{p}$ denote the center and standard deviation of the MF, respectively. Yet, when Gaussian MFs are obtained from training, interpreting the antecedent space can become challenging. We might obtain MFs as in Fig.\ref{fig:FAMvsFAME_L2_LF}(a) or Fig. \ref{fig:FAMvsFAME_L2_LF}(c):
\begin{itemize}
    \item MFs that are not easy to interpret with linguistic terms. 
    \item A partitioning of the universe of discourse that includes numerous MFs, making it challenging to interpret the most significant rule.
\end{itemize}

To address these challenges, the antecedent space is partitioned in a way that enhances interpretability, transforming the FAM into FAME. First of all, we prefer to define the antecedent MFs with two-sided Gaussian (Gauss2MF) as they have more degrees of design flexibility in comparison to the one in \eqref{T1_antecedent_single}. The Gauss2MF is defined as follows:
\begin{equation}
 {\mu_{A_{p}}(z)}= \begin{cases}
    \begin{aligned}
    & \exp \left(-\left(z-c_{p}\right)^{2} / 2 (\sigma_{p}^l)^{2}\right), \text{if } {z} \leq {c_{{p}}} \\
    & \exp \left(-\left(z-c_{p}\right)^{2} / 2 (\sigma_{p}^r)^{2}\right), \text{if } {z} > {c_{{p}}}
    \end{aligned}
 \end{cases}
 \label{gauss2mf}
\end{equation}
where $\sigma_{p}^l$ and $\sigma_{p}^r$ are the left and right standard deviations. Then, to sculpture the antecedent space in an interpretable manner, we parameterize \eqref{gauss2mf} for \( p \in \{1, 2, \dots, P-1\} \) as:
\begin{enumerate}[label=c-\roman*)]
    \item The centers of two consecutive MFs are coupled as:        \begin{equation}\label{calculate_centers}
            c_{p+1} = c_p + 4  \sigma^r_p
        \end{equation}
    Thus, they always satisfy $c_{p+1} \geq c_{p}$ for $\sigma^r_p\geq 0$. 
    \item The standard deviations for \( z\in [c_{p}, c_{p+1}]\) are set as:
  \begin{equation}\label{sigma_conditions}
            \sigma^r_p = \sigma^l_{p+1}
        \end{equation}
Thus, two consecutive MFs share the same \(\sigma\).
\end{enumerate}
This parameterization ensures that the MFs are interpretable as in Fig.\ref{fig:FAMvsFAME_L2_LF}(b) and Fig. \ref{fig:FAMvsFAME_L2_LF}(d). Moreover, for an input $z'$, only two consecutive MFs ($\mu_{A_{p}}(z')$ and $\mu_{A_{p+1}}(z')$) are activated (assuming that for $  |z'|> 4 \sigma^r_p$, $\mu_{A_{p}}(z') \approx 0$). This simplifies the inference in \eqref{y_T1_single} for $z' \in [c_{p^*}, c_{p^*+1}] $ as follows: 
\begin{equation}\label{y_reduced}
g(z';\theta) = \frac{\sum_{p=p^*}^{p^*+1} {\mu_{A_{p}}(z')}\,y_p}{\sum_{p=p^*}^{p^*+1} {\mu_{A_{p}}(z')}} 
\end{equation}
Thus, the resulting inference is highly interpretable since only $P=2$ rules will be activated.

% \subsection{Parametrization of I-T1}

% In this section, we explain the parametrization approach for the Interpretable Type 1 Fuzzy Logic System (I-T1 FLS). Specifically, for the I-T1 FLS, we propose using constrained Gaussian two-membership function (Gauss2MF) fuzzy sets. To enhance interpretability, we impose constraints on the Gauss2MF parameters to ensure that typically only two rules are activated simultaneously. The parameterization is detailed as follows:

% We define the centers of the Gaussian MFs with the constraint:

% \begin{equation}\label{calculate_centers}
% c_{p+1} = c_p + 4  \sigma^r_p
% \end{equation}

% where \( p \in \{1, 2, \dots, P-1\} \) and \( c_1 \) is a scalar representing the initial center. Additionally, we impose the condition:

% \begin{equation}\label{sigma_conditions}
% \sigma^r_p = \sigma^l_{p+1}
% \end{equation}

% Eq. \eqref{calculate_centers} ensure that the distance between any two consecutive centers \( c_p \) and \( c_{p+1} \) is four times the \( \sigma^r_p \) of the \( p \)-th Gaussian MF. Due to the properties of the Gaussian function, this spacing guarantees that, at any given input $(z)$, approximately two rules are activated simultaneously. 

% By implementing these constraints, the I-T1 FLS maintains high interpretability while effectively capturing the necessary input-output relationships through a minimal and manageable set of fuzzy rules.

\section{Learning Framework for FAM/FAME}

Here, we outline the DL framework for FAM and FAME. 
{Algorithm~1}  details the training process for a dataset \( S = \left\{\boldsymbol{x}_{n}, y_{n}\right\}_{n=1}^{N} \), where \( \boldsymbol{x}_n = \left(x_{n, 1}, \ldots, x_{n, D}\right)^{T} \) and \( y_{n} \) while  {Algorithm~2} outlines the Fuzzy Layer computation \footnote{MATLAB implementation. [Online]. Available: \url{https://github.com/gokmenomer/FAME}}. 

To train the FAM/FAME, we first partition the dataset $S$ into \( K \) mini-batches, each containing \( B \) samples. At each epoch, the following optimization problem is minimized:
\begin{equation} \label{loss_cons}
    \min _{\boldsymbol{\theta} \in \mathcal{\boldsymbol{C}}} L_F = \frac{1}{B} \sum_{b=1}^{B} L_2(y_b - \hat{y}_b) + \frac{\lambda}{2} \|\boldsymbol{W}\|_{F}^2,
\end{equation}
where \(L_2\) is the L2 loss. The regularization term, defined by the Frobenius norm \( ||\cdot||_F \), is controlled by the hyperparameter \( \lambda \), and is specifically valid for the PL. The LPs are defined as:
\begin{equation}
\boldsymbol{\theta} = \left\{ \left\{ \boldsymbol{\theta}^{\mathrm{FL}}_i \right\}_{i=1}^{D}, \boldsymbol{\theta}^{\mathrm{PL}} \right\}
\end{equation}
where \( \boldsymbol{\theta}^{\mathrm{PL}} \) represents the LP set of the PL, given by \( \boldsymbol{\theta}^{\mathrm{PL}} = \{\boldsymbol{W}, \boldsymbol{b}\} \), with \( \boldsymbol{W} \in \mathbb{R}^{D \times M} \) and \( \boldsymbol{b} \in \mathbb{R}^{D \times 1} \), while LP set 
\begin{itemize} 
     \item For FAM: $\boldsymbol{\theta}^{\mathrm{FL}}$ = $\{\boldsymbol{c}, \boldsymbol{\sigma}, \boldsymbol{a}, \boldsymbol{a}_0\}$, where $\boldsymbol{c}, \boldsymbol{\sigma}, \boldsymbol{a}, \boldsymbol{a}_0  \in \mathbb{R}^{P \times 1}$.
    \item For FAME: $\boldsymbol{\theta}^{\mathrm{FL}}$ = $\{c_1, \sigma^l_1, \boldsymbol{\sigma^r}, \boldsymbol{a}, \boldsymbol{a}_0\}$, where $c_1$ and $\sigma^l_1$ are scalar LPs and $\boldsymbol{\sigma^r}, \boldsymbol{a}, \boldsymbol{a}_0  \in \mathbb{R}^{P \times 1}$.
\end{itemize}
% In summary, the FAME includes a total of \(D(3P + M + 3)\) LPs while FAM \textcolor{red}{XXX}.

During learning, we must ensure $\sigma_p>0, \forall p$, thus \(\theta \in \mathcal{\boldsymbol{C}}\) must hold. Especially for FAME, this constraint becomes particularly crucial since \( c_p \) is defined as in c-i). Given that DL optimizers are unconstrained techniques, we deploy the following parameterization trick like in \cite{guven,koklu1}:  
\begin{equation}
    \sigma_p = abs(\sigma'_p)
\end{equation}
that transforms \eqref{loss_cons} into an unconstrained one through new LPs $\sigma_p' \in [-\infty, \infty], \forall p $.

\section{Performance Analysis} \label{performance}
This section provides a comprehensive analysis of the learning performance through a dual-fold evaluation.
        \begin{enumerate}[label=(\roman*)]
            \item Analyzing the impact of feature space on the performance of FAM and FAME, both quantitatively (accuracy) and qualitatively (the antecedent space).
            \item Evaluating the performance of FAM and FAME against their MFLS counterparts.
        \end{enumerate}

\begin{algorithm} [t]
\caption{DL-based FAME Training Algorithm}
\begin{algorithmic}[1] % [1] ensures line numbering starts and increases sequentially
\label{alg:gam_training}
\State \textbf{Input:} $N$ training samples $(x_{n},y_{n})^{N}_{n=1}$
\State $P$, number of rules
\State $mbs$, mini-batch size
\State \textbf{Output:} LP set $\boldsymbol{\theta}$
\State Initialize $\boldsymbol{\theta}$
\For{\textbf{each } $mbs$ in $N$} 
    \State $\boldsymbol{z} \gets \text{ProjectionLayer}(\boldsymbol{x}; \boldsymbol{\theta}^{\mathrm{PL}})$ \Comment{Eq. \eqref{projection_layer}}
    \vspace{0.03cm}
    \State $\hat{Y} \gets \mathrm{FuzzyLayer}(\boldsymbol{z}; \{\boldsymbol{\theta}^{\mathrm{FL}}_i \}_{i=1}^{D})$
    \State $\hat{Y} \gets \sum_{d=1}^{D} \hat{Y}_d$
    \State Compute $L$ \Comment{Eq. \eqref{loss_cons}}
    \State Compute the gradient ${\partial L}/{\partial {\theta}}$
    \State Update ${\theta}$ via Adam optimizer
\EndFor
\State $\theta = \arg \min L$
\State \textbf{Return} $\theta$
\end{algorithmic}
\end{algorithm}

\begin{algorithm} [t]
\caption{FuzzyLayer Computation}
\begin{algorithmic}[1]
\label{alg:T1_gam}
\State \textbf{Input:} $\boldsymbol{z}, \{\boldsymbol{\theta}^{\mathrm{FL}}_i \}_{i=1}^{D}$
\State \textbf{Output:} $\hat{Y}$ 
\State Initialize $\hat{Y} \gets \varnothing$ 
\For{\textbf{each } $d$ in $D$}
    \State $\hat{y}_d \gets \mathrm{FLS}(\boldsymbol{z}_d; \boldsymbol{\theta}^{\mathrm{FL}}_d)$
    \vspace{0.03cm}
    \State $\hat{Y} \gets \hat{Y} \cup \{\hat{y}_d \}$
\EndFor
\State \textbf{Return} $\hat{Y}$
\end{algorithmic}
\end{algorithm}

\subsection{Design of Experiments}
We evaluate the RMSE performance of FLSs on benchmark datasets, including Abalone (ABA), AIDS, Boston Housing (BH), Parkinson Motor UPDRS (PM), White Wine (WW), and Concrete Strength (CS). All datasets are preprocessed using z-score normalization, with 70\% of the data allocated to the training set and 30\% to the test set.

The learning performances of the following FAMs/FAMEs are examined:
\begin{itemize}
    \item \textbf{FAM}, which uses reduced input space $(z_i \in \boldsymbol{Z})$, as introduced in Section \ref{FAM_intro}.
    \item \textbf{Vanilla FAM}  (V-FAM), is FAM without a PL, i.e., the FL processes $x_i \in \boldsymbol{X}$.
    \item \textbf{FAME}, which uses $z_i \in \boldsymbol{Z}$, as presented in Section \ref{Gauss2MF_Intro}.
    \item \textbf{Vanilla FAME}(V-FAME), is FAME without a PL, i.e., the FL processes $x_i \in \boldsymbol{X}$.    
    \end{itemize}
We also train MFLSs using Gaussian MFs and MFLSE with Gauss2MFs in the antecedent as parametrized within the paper.    
    \begin{itemize}
        \item \textbf{Vanilla MFLS/MFLSE}(V-MFLS/MFLSE), processes $x_i \in\boldsymbol{X}$ with a rule base as defined in \cite{infus}.
        \item \textbf{CDR-MFLS/MFLSE} \cite{cdr_dr}, which uses $z_i \in \boldsymbol{Z}$, integrates the PL with MFLS/MFLSE.
        \item \textbf{DR-MFLS/MFLSE} \cite{cdr_dr}, maintains the same antecedent structure as CDR-MFLS/MFLSE but utilizes $x_i \in \boldsymbol{X}$ in the consequent part. 
    %     Specifically, for each $p$,
    % \begin{equation}\label{DR_consequent}
    %     y_{p} = \sum\limits_{m=1}^{M} a_{p,m} \, x_{m} + a_{p,0}.
    % \end{equation} 
\end{itemize}
Table \ref{tab:parameter-sets} summarizes the LP set size of the defined FLSs.

\begin{table}[b]
    \centering
    \caption{\#LPs of handled FLSs}
    \label{tab:parameter-sets}
    \begin{tabular}{lc}
        \hline
        \textbf{Models} & \textbf{\#LP} \\
        \hline
        FAM& $4PD + (M+1)D$ \\
        V-FAM& $4PD$ \\
        FAME & $D(3P + 2) + (M+1)D$ \\
        V-FAME & $D(3P + 2)$ \\
        V-MFLS & $P(3M + 1)$ \\
        CDR-MFLS & $P(3D + 1) + (M+1)D$ \\
        DR-MFLS & $P(2D + M + 1) + (M+1)D$ \\
        V-MFLSE & $M(2P + 2) + P$ \\
        CDR-MFLSE & $D(2P + 2) + P + (M+1)D$ \\
        DR-MFLSE & $P(D + M + 1) + 2D + (M+1)D$ \\
        \hline
    \end{tabular}   
\end{table}

\begin{figure*}[ht]
        \centering
        \subfigure[FAM(4): $L_2$]
        {
        \includegraphics[width=0.39\textwidth]
                    {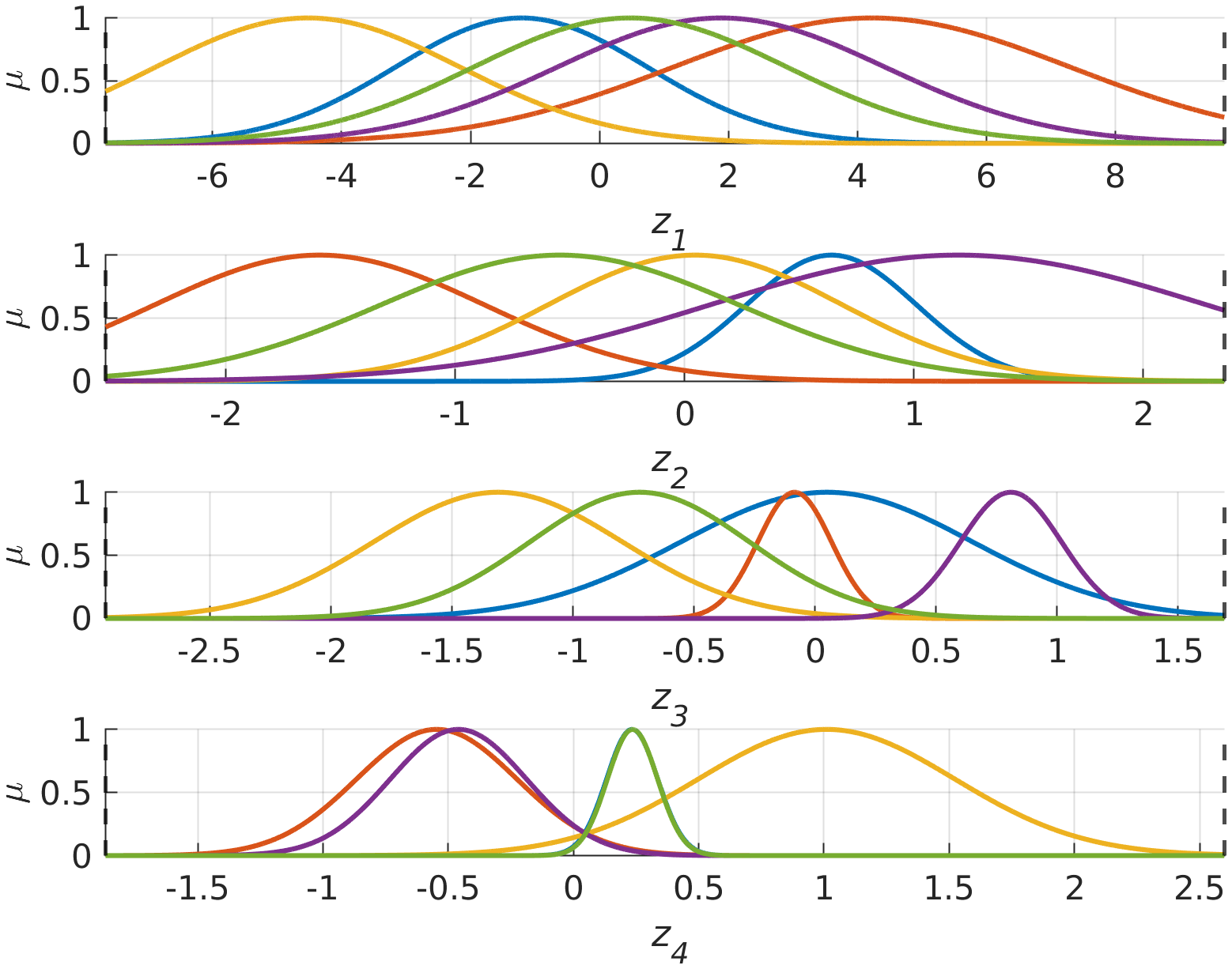}
                    % \caption{IT2-FLS-H}
                    \label{fig:FAM(4)_antacadents}
        }
        \subfigure[FAME(4): $L_2$]
        {
        \includegraphics[width=0.39\textwidth]
                    {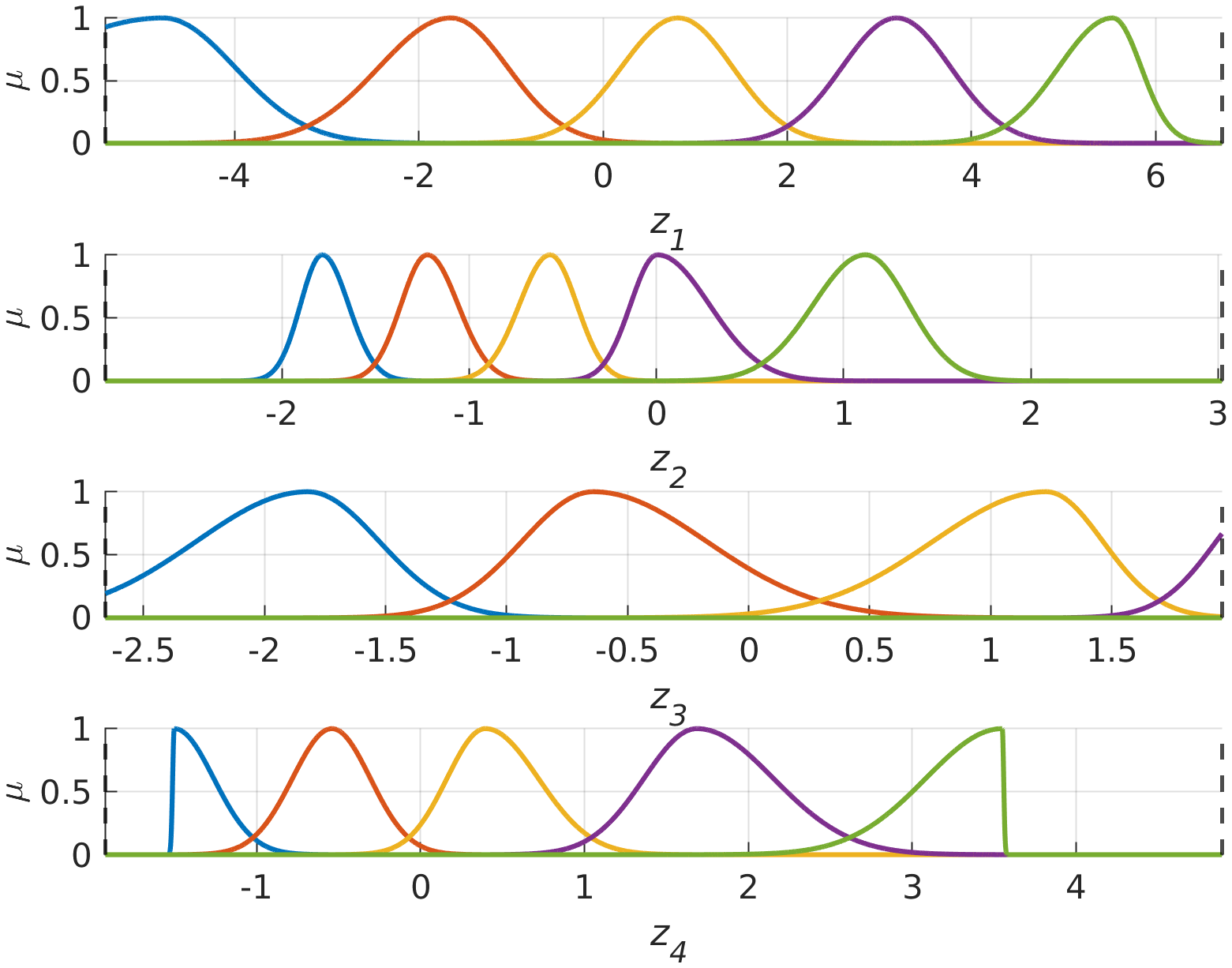}
                    % \caption{IT2-FLS-HS}
                    \label{fig:FAME(4)_antacadents}
        
        }

        \subfigure[FAM(4): $L_F$]
        {
        \includegraphics[width=0.39\textwidth]
                    {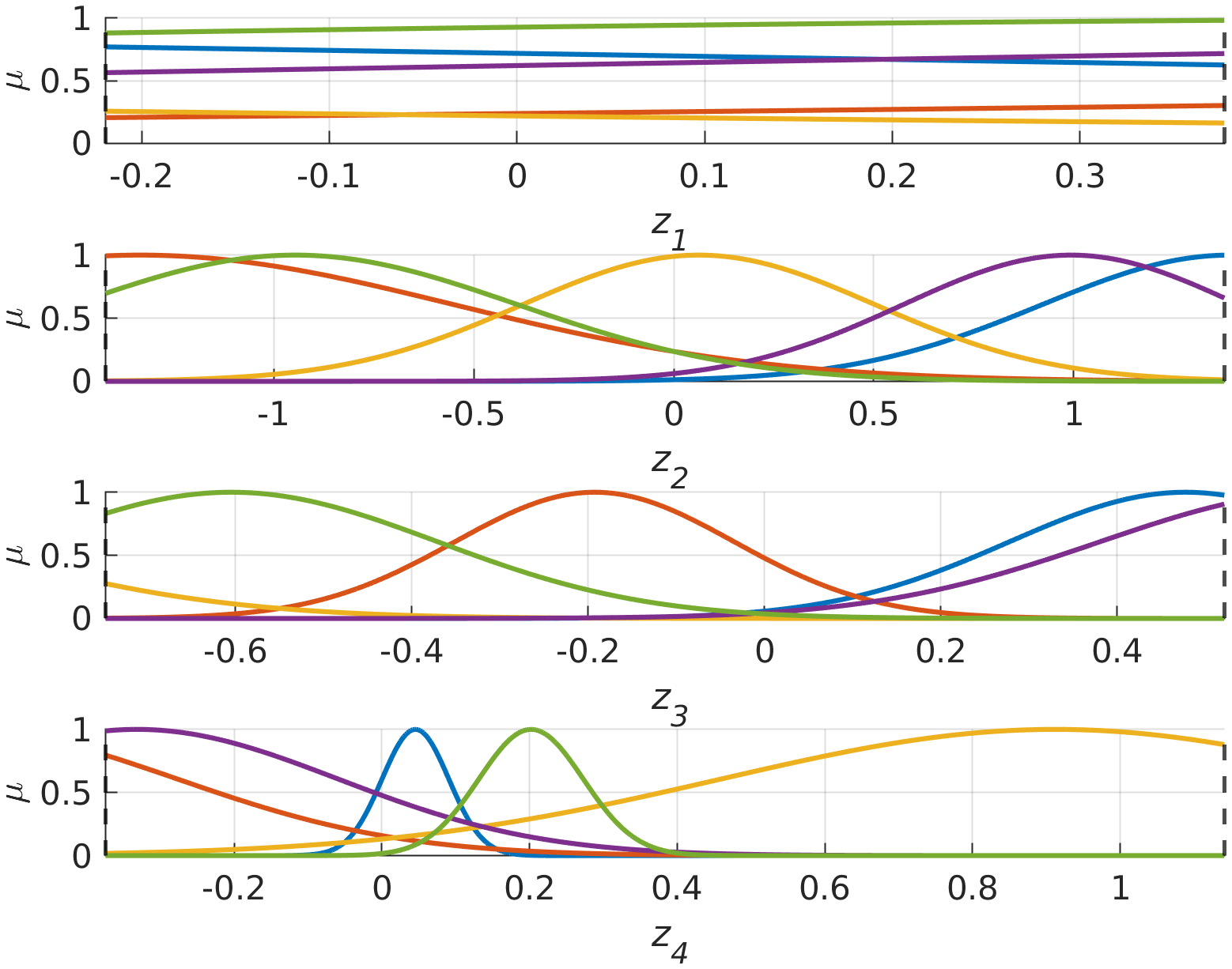}
                    % \caption{IT2-FLS-H}
                    \label{fig:FAM(4)_antacadents_lf}
        }
        \subfigure[FAME(4): $L_F$]
        {
        \includegraphics[width=0.39\textwidth]
                    {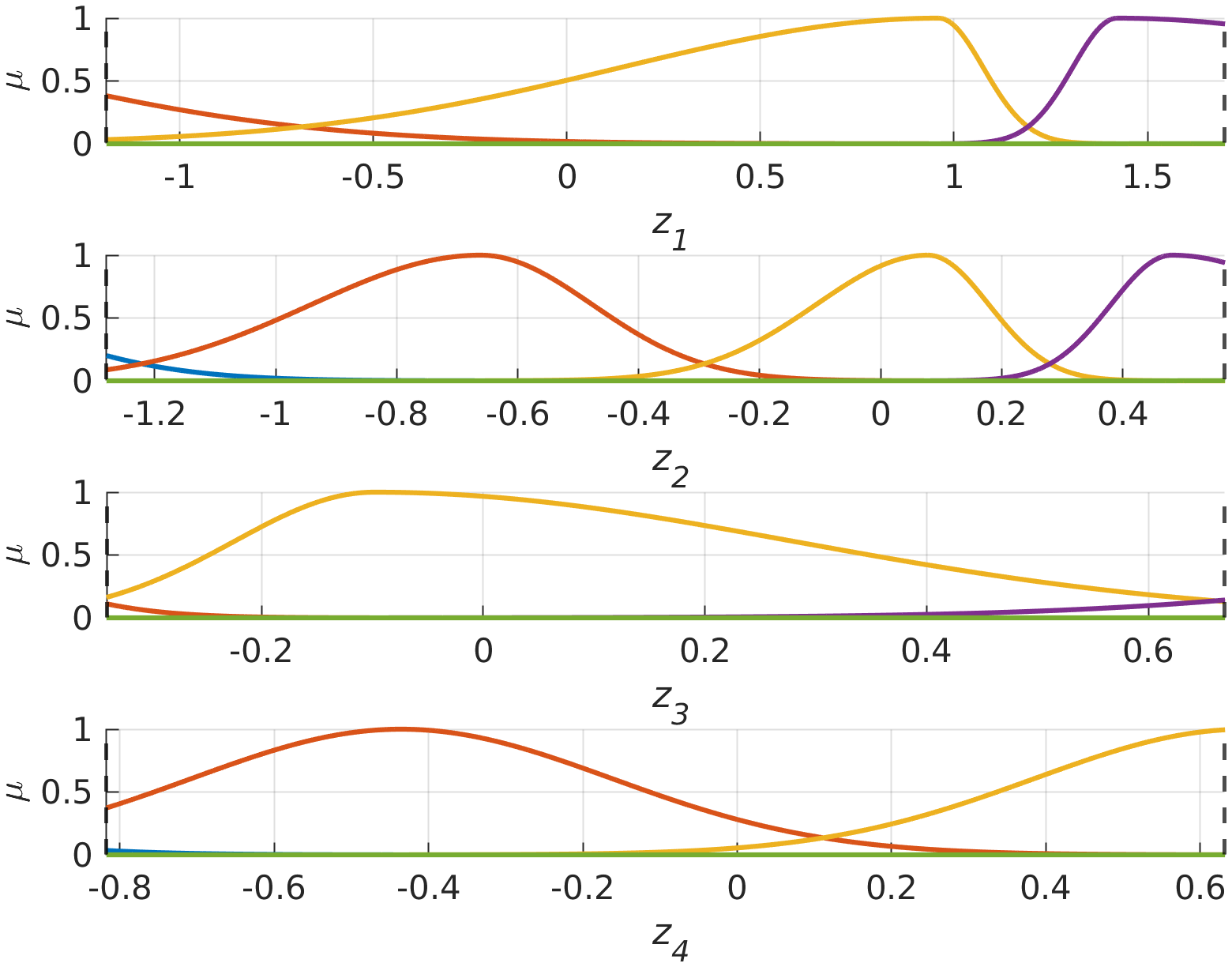}
                    % \caption{IT2-FLS-HS}
                    \label{fig:FAME(4)_antacadents_lf}
        
        }
        \caption{Visualization of MFs in the effective universe of discourse, defined by the max and min values of learned \(z_i\)'s, and indicated by dashed lines.}

          \label{fig:FAMvsFAME_L2_LF}
\end{figure*}

We trained all FLSs with $P = 5$ rules and used the same learning rate of $0.01$, a regularization parameter value $\lambda = 0.05$, and mini-batch size (mbs) of $64$ for $100$ epochs. For the PM dataset, training was extended to $1000$ epochs with $mbs=512$. FLSs with a PL were trained using both $L_2$ and $L_F$ losses with \( D = \{2, 4, 8\} \), while those without a PL were trained with only the $L_2$ loss. All experiments were conducted in MATLAB\textsuperscript{\textregistered} and repeated with 10 different initial seeds for statistical analysis.

\begin{table*}[t] % Use table* for spanning across two columns
\centering
\begin{minipage}{0.48\textwidth} % First column: adjust width as needed
    \centering
    \caption{Performance Analysis of FAM with $L_2$ over 10 Experiments}
    \resizebox{\columnwidth}{!}{%
    \begin{tabular}{lcccccc}
    \hline
    Dataset & Metric & V-MFLS & V-FAM & FAM(2) & FAM(4) & FAM(8) \\
    \hline
\multirow{2}{*}{ABA} 
  & \#LP & 125 & 160 & 58 & 116 & 232 \\
  & RMSE & 67.24($\pm$2.08) & 67.07($\pm$1.72) & 66.01($\pm$1.86) & \textbf{65.83($\pm$1.81)} & 66.07($\pm$1.75) \\
\hline
\multirow{2}{*}{AIDS} 
  & \#LP & 350 & 460 & 88 & 176 & 352 \\
  & RMSE & 70.16($\pm$1.36) & 70.72($\pm$2.08) & \textbf{67.61($\pm$1.86)} & 69.35($\pm$1.71) & 73.08($\pm$3.56) \\
\hline
\multirow{2}{*}{BH} 
  & \#LP & 200 & 260 & 68 & 136 & 272 \\
  & RMSE & 42.01($\pm$2.71) & 41.53($\pm$4.20) & \textbf{41.44($\pm$6.74)} & 42.93($\pm$4.61) & 42.27($\pm$5.51) \\
\hline
\multirow{2}{*}{PM} 
  & \#LP & 290 & 380 & 80 & 160 & 320 \\
  & RMSE & 65.00($\pm$3.93) & 79.72($\pm$2.13) & 71.00($\pm$2.62) & 52.69($\pm$5.12) & \textbf{39.50($\pm$4.81)} \\
\hline
\multirow{2}{*}{WW} 
  & \#LP & 170 & 220 & 64 & 128 & 256 \\
  & RMSE & \textbf{80.61($\pm$1.55)} & 80.67($\pm$1.43) & 83.06($\pm$1.84) & 81.15($\pm$2.60) & 81.70($\pm$4.00) \\
\hline
\multirow{2}{*}{CS} 
  & \#LP & 125 & 160 & 58 & 116 & 232 \\
  & RMSE & 36.01($\pm$2.48) & 36.92($\pm$1.27) & 43.89($\pm$1.28) & 38.49($\pm$1.76) & \textbf{34.49($\pm$1.64)} \\
  \hline
    \end{tabular}}
    \label{tab:fam-results_l2}
\end{minipage}
\hfill
\begin{minipage}{0.48\textwidth} % Second column: adjust width as needed
    \centering
    \caption{Performance Analysis of FAM with $L_F$ Over 10 Experiments}
    \resizebox{\columnwidth}{!}{%
    \begin{tabular}{lcccccc}
    \hline
    Dataset & Metric & V-MFLS & V-FAM & FAM(2) & FAM(4) & FAM(8) \\
    \hline
\multirow{2}{*}{ABA} 
  & \#LP & 125 & 160 & 58 & 116 & 232 \\
  & RMSE & 67.24($\pm$2.08) & 67.07($\pm$1.72) & \textbf{65.67($\pm$1.70)} & 65.76($\pm$1.68) & 65.73($\pm$1.64) \\
\hline
\multirow{2}{*}{AIDS} 
  & \#LP & 350 & 460 & 88 & 176 & 352 \\
  & RMSE & 70.16($\pm$1.36) & 70.72($\pm$2.08) & \textbf{67.65($\pm$2.24)} & 69.13($\pm$1.96) & 70.77($\pm$2.84) \\
\hline
\multirow{2}{*}{BH} 
  & \#LP & 200 & 260 & 68 & 136 & 272 \\
  & RMSE & 42.01($\pm$2.71) & 41.53($\pm$4.20) & 43.64($\pm$5.40) & 41.26($\pm$3.67) & \textbf{40.25($\pm$6.60)} \\
\hline
\multirow{2}{*}{PM} 
  & \#LP & 290 & 380 & 80 & 160 & 320 \\
  & RMSE & 65.00($\pm$3.93) & 79.72($\pm$2.13) & 73.07($\pm$1.99) & 63.10($\pm$2.70) & \textbf{49.28($\pm$2.70)} \\
\hline
\multirow{2}{*}{WW} 
  & \#LP & 170 & 220 & 64 & 128 & 256 \\
  & RMSE & 80.61($\pm$1.55) & 80.67($\pm$1.43) & 82.72($\pm$1.84) & 80.47($\pm$1.46) & \textbf{79.96($\pm$1.04)} \\
\hline
\multirow{2}{*}{CS} 
  & \#LP & 125 & 160 & 58 & 116 & 232 \\
  & RMSE & \textbf{36.01($\pm$2.48)} & 36.92($\pm$1.27) & 42.79($\pm$1.94) & 39.38($\pm$1.45) & 36.99($\pm$2.83) \\
\hline
    \end{tabular}}
    \label{tab:fam-results_l2f}
\end{minipage}
\vspace{0.5cm} % Add vertical space between rows
\begin{minipage}{0.48\textwidth} % Third table below first
    \centering
    \caption{Performance Analysis of FAME with $L_2$ Over 10 Experiments}
    \resizebox{\columnwidth}{!}{%
    \begin{tabular}{lcccccc}
    \hline
    Dataset & Metric & V-MFLSE & V-FAME & FAME(2) & FAME(4) & FAME(8) \\
   \hline
\multirow{2}{*}{ABA} 
  & \#LP & 101 & 136 & 52 & 104 & 208 \\
  & RMSE & 67.25($\pm$2.24) & 68.40($\pm$1.98) & 66.30($\pm$2.05) & \textbf{65.31($\pm$2.05)} & 65.44($\pm$1.37) \\
\hline
\multirow{2}{*}{AIDS} 
  & \#LP & 281 & 391 & 82 & 164 & 328 \\
  & RMSE & \textbf{70.52($\pm$2.20)} & 71.21($\pm$2.27) & 71.99($\pm$3.71) & 71.15($\pm$2.11) & 74.57($\pm$2.60) \\
\hline
\multirow{2}{*}{BH} 
  & \#LP & 161 & 221 & 62 & 124 & 248 \\
  & RMSE & 43.37($\pm$5.32) & \textbf{43.08($\pm$4.17)} & 44.90($\pm$5.96) & 46.33($\pm$3.32) & 45.60($\pm$6.42) \\
\hline
\multirow{2}{*}{PM} 
  & \#LP & 233 & 323 & 74 & 148 & 296 \\
  & RMSE & 79.13($\pm$3.46) & 81.01($\pm$1.61) & 77.34($\pm$2.36) & 61.60($\pm$2.73) & \textbf{44.79($\pm$4.93)} \\
\hline
\multirow{2}{*}{WW} 
  & \#LP & 137 & 187 & 58 & 116 & 232 \\
  & RMSE & 82.32($\pm$1.33) & 81.13($\pm$1.60) & 83.24($\pm$1.96) & \textbf{80.69($\pm$1.44)} & 81.22($\pm$2.15) \\
\hline
\multirow{2}{*}{CS} 
  & \#LP & 101 & 136 & 52 & 104 & 208 \\
  & RMSE & 43.75($\pm$4.31) & \textbf{38.88($\pm$1.61)} & 58.73($\pm$14.04) & 41.79($\pm$2.88) & 38.98($\pm$3.43) \\
\hline
    \end{tabular}}
    \label{tab:fame-results_l2}
\end{minipage}
\hfill
\begin{minipage}{0.48\textwidth} % Fourth table below second
    \centering
    \caption{Performance Analysis of FAME with $L_F$ Over 10 Experiments}
    \resizebox{\columnwidth}{!}{%
    \begin{tabular}{lcccccc}
    \hline
    Dataset & Metric & V-MFLSE & V-FAME & FAME(2) & FAME(4) & FAME(8) \\
   \hline
\multirow{2}{*}{ABA} 
  & \#LP & 101 & 136 & 52 & 104 & 208 \\
  & RMSE & 67.25($\pm$2.24) & 68.40($\pm$1.98) & 66.55($\pm$2.16) & \textbf{65.44($\pm$1.85)} & 65.74($\pm$2.12) \\
\hline
\multirow{2}{*}{AIDS} 
  & \#LP & 281 & 391 & 82 & 164 & 328 \\
  & RMSE & 70.52($\pm$2.20) & 71.21($\pm$2.27) & 70.78($\pm$4.34) & \textbf{68.11($\pm$2.05)} & 69.59($\pm$2.28) \\
\hline
\multirow{2}{*}{BH} 
  & \#LP & 161 & 221 & 62 & 124 & 248 \\
  & RMSE & 43.37($\pm$5.32) & 43.08($\pm$4.17) & 43.96($\pm$6.44) & 42.76($\pm$6.17) & \textbf{39.66($\pm$4.85)} \\
\hline
\multirow{2}{*}{PM} 
  & \#LP & 233 & 323 & 74 & 148 & 296 \\
  & RMSE & 79.13($\pm$3.46) & 81.01($\pm$1.61) & 77.78($\pm$2.29) & 64.04($\pm$5.91) & \textbf{57.28($\pm$4.17)} \\
\hline
\multirow{2}{*}{WW} 
  & \#LP & 137 & 187 & 58 & 116 & 232 \\
  & RMSE & 82.32($\pm$1.33) & 81.13($\pm$1.60) & 82.97($\pm$1.58) & 80.91($\pm$1.23) & \textbf{80.76($\pm$1.53)} \\
\hline
\multirow{2}{*}{CS} 
  & \#LP & 101 & 136 & 52 & 104 & 208 \\
  & RMSE & 43.75($\pm$4.31) & 38.88($\pm$1.61) & 51.30($\pm$9.43) & 41.19($\pm$1.98) & \textbf{36.67($\pm$2.07)} \\
\hline
    \end{tabular}}
    \label{tab:fame-results_l2f}
\end{minipage}

\begin{threeparttable} % To add tablenotes
\vspace{-0.5cm}
    \begin{tablenotes}
        \item (1) RMSE values are scaled by 100.
        \item (2) Measures that are highlighted indicate the best performance.
    \end{tablenotes}
\end{threeparttable}
\vspace{-0.6cm}
\end{table*}

\subsection{Impact of Feature Space on FAM and FAME Performance}

Table \ref{tab:fam-results_l2}-\ref{tab:fame-results_l2f} summarizes RMSE values of FAM and FAME over 10 experiments for both the deployment of $L_2$ and $L_f$ alongside their \#LP. Here, FAM and FAME with \( D = \{2, 4, 8\} \) are referred to as \(\text{FAM}(D)\) or \(\text{FAME}(D)\), respectively. To define a baseline performance, we also included the results of V-MFLS. We observe that: 
\begin{itemize}
    \item V-FAM and V-FAME, which are additive models, perform similarly to V-MFLS and V-MFLSE. FAM(\(D\)) and FAME(\(D\)), additive models equipped with a PL, outperform them. Yet, the setting of \(D\) depends on the dataset, emphasizing its importance as a key hyperparameter.
    \item FAM typically achieves better performance than FAME. This outcome is anticipated because FAME's reduced number of LPs limits its learning capacity, often resulting in lower performance compared to FAM.
    \item FAM and FAME exhibit comparable performance regardless of whether \(L_2\) or \(L_F\) is used.

\end{itemize}

Now, let us examine how the antecedent space is shaped after training. Fig. \ref{fig:FAMvsFAME_L2_LF} shows the learned MFs from an ABA experiment, where FAM achieved a slightly better RMSE value. Here, we plot the effective universe of discourse of the antecedent MFs, defined as the range of $z_i$ obtained after training (i.e., max and min values). Observe that: 
\begin{itemize} 

    % \item In Fig.~\ref{fig:FAMvsFAME_L2_LF}(a), MFs overlap, making it difficult to assign distinct linguistic labels (e.g., very cold, $\ldots$, very hot). For instance, around \(z_1 = 0\), at least 3 MFs have similar membership grades. In contrast, Fig.~\ref{fig:FAMvsFAME_L2_LF}(b) uses at most two MFs at any point \(z_i\), allowing clearer distinctions among linguistic terms.

    % \item  While FAME does not consistently outperform FAM and may be less accurate on certain datasets, it offers greater transparency. This is achieved through the use of Gauss2MF in the antecedents, which enhances the interpretability of the system at a slight cost to accuracy.

    \item In Fig.~\ref{fig:FAMvsFAME_L2_LF}(a), the MFs of FAM overlap, making it difficult to assign distinct linguistic labels. For example, around \(z_1 = 0\), multiple MFs could share the same label. In contrast, FAME shows improved interpretability by limiting the active MFs to at most two for any $z_i$ making it easier to label the MFs, as shown in Fig.~\ref{fig:FAMvsFAME_L2_LF}(b). We also observe input spaces defined by one MF in FAME where a single rule is learned to represent this input range. This simplifies the model and improves interpretability by reducing the complexity of the rules learned.

    \item The impact of the loss is evident in Fig.~\ref{fig:FAMvsFAME_L2_LF}, where \(L_F\) results in a narrower effective universe of discourse compared to \(L_2\). Fig.~\ref{fig:FAMvsFAME_L2_LF}(c) highlights the challenge of interpreting the FAM as all MFs overlap. Unlike for FAME, we observe from Fig.~\ref{fig:FAMvsFAME_L2_LF}(d) that the number of interpretable MFs within effective input space is reduced, and thus fewer rules are needed for explanation.  
   
    % \item   The impact of loss functions is evident in Fig.~\ref{fig:FAMvsFAME_L2_LF}, where \(L_F\) yields a narrower effective universe of discourse than \(L_2\). In particular, Fig.~\ref{fig:FAMvsFAME_L2_LF}(c) shows how it is difficult to interpret the antecedent part of \(z_1\) since all MFs overlap around the effective universe of discourse. On the other hand, we can clearly say that generally number of effective rule decreases, also for $z_3$, we only need 1 rule to explain the effective universe of discourse as in Fig.~\ref{fig:FAMvsFAME_L2_LF}(d).

    % \item \textcolor{red}{şekiller üzerinden yorum yazalım, L2'ye kıyasla L2F yapınca effektif universe of discourse küçülüyor. FAM(4) z1 ne kadar anlamasız dedik! FAME'de efektif uzay küçüldüğü için kural azalıyor vurgusu lazım} 
    % \item  Both \(L_2\) and \(L_F\) losses result in similar overall performance, with \(L_F\) leading to more selective rule activations and a narrower input range, without affecting the overall efficacy of the model.

\end{itemize}

In conclusion, the choice between FAM and FAME depends on the balance between explainability and performance. Yet, FAME with \(L_F\) improves interpretability by reducing the number of rules and offering clearer linguistic labels for antecedents when compared to FAM  at a slight cost to accuracy.

\subsection{Assessing the Performance of FAM and FAME vs MFLS}

In Table \ref{tab:bestofall_l2} and Table \ref{tab:bestofall_lf}, we show the RMSE values for all trained FLSs over 10 experiments using $L_2$ and $L_F$ loss functions, respectively. Here, we only reported the results of FLSs with a PL (FAM, FAME, CDR, and DR) with their best hyperparameter setting of \( D \) selected from \( D = \{2, 4, 8\} \). We also provided the rankings over the handled 6 datasets for an easy comparison for each model. We can observe that:

\begin{itemize}

    \item FAM/FAME generally exhibits competitive performances. Notably, FAM shows the best overall performance according to average rankings. 
    \item FAME is a robust performer as it is in the top 3 in both average ranks while also resulting in interpretability and a manageable active set of rules. 
    \item  FAM/FAME has a better performance compared to its vanilla counterparts. This shows PL has a positive effect.
  % \item Compared to MFLSs, additive models with SFLS yield slightly better RMSE performance and average ranking. \textcolor{red}{neden additive yazdık?-ranking yazdığımızı yukarıda değinmemiz lazım}
  % \item Including PL consistently improves model performance. \textcolor{red}{çok genel}
  % \item Using the $L_F$ instead of only $L_2$ further enhances the performance of models that include PL. In particular, the proposed FAM and FAME benefit more from $L_F$, demonstrating that PL indeed provides additional gains over their vanilla counterparts.
\end{itemize}

% Overall, FAM and FAME consistently demonstrate superior interpretability compared to other MFLS-based approaches. The inclusion of the PL further enhances each FLS configuration by reducing dimensionality and improving membership function utilization. Additionally, the \(L_F\) loss helps narrow the effective input range, prompting more selective activation of rules and MFs. These combined benefits yield robust, efficient, and interpretable fuzzy inference solutions across diverse datasets.
In summary, we conclude that FAM and FAME do not result in accuracy decrements compared to their MFLS counterparts, despite being based on SFLSs, which also process input interactions. FAME, in particular, is generally a robust performer across all datasets, with the added benefit of interpretability. 
\begin{table*}[ht]
\centering
\caption{ Performance Analysis of Best of All with $L_2$ Over 10 Experiments}
\scriptsize               
\setlength{\tabcolsep}{2pt} 
\begin{threeparttable}
\resizebox{\textwidth}{!}{%
\begin{tabular}{lccccccccccc}
\hline
\textbf{Dataset} & \textbf{Metric} & \textbf{FAM} & \textbf{V-FAM} & \textbf{FAME} & \textbf{V-FAME} & \textbf{V-MFLS} & \textbf{V-MFLSE} & \textbf{CDR-MFLS} & \textbf{CDR-MFLSE} & \textbf{DR-MFLS} & \textbf{DR-MFLSE} \\
\hline
\multirow{2}{*}{ABA} 
  & \#LP & 116 & 160 & 104 & 136 & 125 & 101 & 53 & 173 & 83 & 173 \\
  & RMSE & 65.83($\pm$1.81) & 67.07($\pm$1.72) & \textbf{65.31($\pm$2.05)} & 68.40($\pm$1.98) & 67.24($\pm$2.08) & 67.25($\pm$2.24) & 66.39($\pm$1.82) & 66.22($\pm$1.54) & 66.73($\pm$1.91) & 66.17($\pm$1.59) \\
\hline
\multirow{2}{*}{AIDS} 
  & \#LP & 88 & 460 & 164 & 391 & 350 & 281 & 83 & 77 & 188 & 182 \\
  & RMSE & 67.61($\pm$1.86) & 70.72($\pm$2.08) & 71.15($\pm$2.11) & 71.21($\pm$2.27) & 70.16($\pm$1.36) & 70.52($\pm$2.20) & \textbf{65.97($\pm$2.81)} & 69.25($\pm$3.08) & 71.52($\pm$2.29) & 71.40($\pm$2.65) \\
\hline
\multirow{2}{*}{BH} 
  & \#LP & 68 & 260 & 62 & 221 & 200 & 161 & 63 & 57 & 118 & 238 \\
  & RMSE & \textbf{41.44($\pm$6.74)} & 41.53($\pm$4.20) & 44.90($\pm$5.96) & 43.08($\pm$4.17) & 42.01($\pm$2.71) & 43.37($\pm$5.32) & 43.54($\pm$5.25) & 46.81($\pm$6.75) & 42.37($\pm$6.14) & 44.09($\pm$6.03) \\
\hline
\multirow{2}{*}{PM} 
  & \#LP & 320 & 380 & 296 & 323 & 290 & 233 & 285 & 261 & 340 & 316 \\
  & RMSE & \textbf{39.50($\pm$4.81)} & 79.72($\pm$2.13) & 44.79($\pm$4.93) & 81.01($\pm$1.61) & 65.00($\pm$3.93) & 79.13($\pm$3.46) & 46.45($\pm$3.91) & 56.33($\pm$5.99) & 47.42($\pm$6.25) & 53.60($\pm$2.83) \\
\hline
\multirow{2}{*}{WW} 
  & \#LP & 128 & 220 & 116 & 187 & 170 & 137 & 113 & 197 & 104 & 212 \\
  & RMSE & 81.15($\pm$2.60) & 80.67($\pm$1.43) & 80.69($\pm$1.44) & 81.13($\pm$1.60) & \textbf{80.61($\pm$1.55)} & 82.32($\pm$1.33) & 81.70($\pm$2.43) & 81.17($\pm$1.94) & 81.05($\pm$1.66) & 80.90($\pm$2.15) \\
\hline
\multirow{2}{*}{CS} 
  & \#LP & 232 & 160 & 208 & 136 & 125 & 101 & 197 & 173 & 197 & 173 \\
  & RMSE & \textbf{34.49($\pm$1.64)} & 36.92($\pm$1.27) & 38.98($\pm$3.43) & 38.88($\pm$1.61) & 36.01($\pm$2.48) & 43.75($\pm$4.31) & 41.35($\pm$3.90) & 44.18($\pm$3.75) & 40.78($\pm$3.80) & 40.64($\pm$2.63) \\
\hline
\hline
Average Rank & & \textbf{2.34} & 4.84 & 4.50 & 7.17 & 4.17 & 7.84 & 5.50 & 6.84 & 6.00 & 5.84 \\
\hline
\end{tabular}
}% end \resizebox
% \begin{tablenotes}
% % \item[(1)] RMSE values are scaled by 100.
% % \item[(2)] The bold values are the best results in each row.
% \end{tablenotes}
\end{threeparttable}
\label{tab:bestofall_l2}
\end{table*}

%%%%% Best of All L2F
\begin{table*}[t]
\vspace{-0.2cm}
\centering
\caption{ Performance Analysis of Best of All with $L_F$ Over 10 Experiments}
\scriptsize                % Use smaller font
\setlength{\tabcolsep}{2pt} % Tighten column spacing
\begin{threeparttable}
\resizebox{\textwidth}{!}{%
\begin{tabular}{lccccccccccc}
\hline
\textbf{Dataset} & \textbf{Metric} & \textbf{FAM} & \textbf{V-FAM} & \textbf{FAME} & \textbf{V-FAME} & \textbf{V-MFLS} & \textbf{V-MFLSE} & \textbf{CDR-MFLS} & \textbf{CDR-MFLSE} & \textbf{DR-MFLS} & \textbf{DR-MFLSE} \\
\hline
\multirow{2}{*}{ABA} 
  & \#LP & 58 & 160 & 104 & 136 & 125 & 101 & 101 & 89 & 121 & 173 \\
  & RMSE & 65.67($\pm$1.70) & 67.07($\pm$1.72) & \textbf{65.44($\pm$1.85)} & 68.40($\pm$1.98) & 67.24($\pm$2.08) & 67.25($\pm$2.24) & 65.60($\pm$1.76) & 65.49($\pm$1.87) & 66.12($\pm$1.72) & 65.68($\pm$1.82) \\
\hline
\multirow{2}{*}{AIDS} 
  & \#LP & 88 & 460 & 164 & 391 & 350 & 281 & 83 & 77 & 188 & 182 \\
  & RMSE & 67.65($\pm$2.24) & 70.72($\pm$2.08) & 68.11($\pm$2.05) & 71.21($\pm$2.27) & 70.16($\pm$1.36) & 70.52($\pm$2.20) & \textbf{66.31($\pm$1.52)} & 67.83($\pm$2.11) & 69.28($\pm$2.36) & 70.06($\pm$2.34) \\
\hline
\multirow{2}{*}{BH} 
  & \#LP & 272 & 260 & 248 & 221 & 200 & 161 & 121 & 213 & 262 & 238 \\
  & RMSE & 40.25($\pm$6.60) & 41.53($\pm$4.20) & 39.66($\pm$4.85) & 43.08($\pm$4.17) & 42.01($\pm$2.71) & 43.37($\pm$5.32) & 41.25($\pm$6.23) & 42.62($\pm$6.60) & 41.11($\pm$6.08) & \textbf{38.42($\pm$4.90)} \\
\hline
\multirow{2}{*}{PM} 
  & \#LP & 320 & 380 & 296 & 323 & 290 & 233 & 285 & 261 & 340 & 316 \\
  & RMSE & 49.28($\pm$2.70) & 79.72($\pm$2.13) & 57.28($\pm$4.17) & 81.01($\pm$1.61) & 65.00($\pm$3.93) & 79.13($\pm$3.46) & \textbf{44.94($\pm$5.39)} & 55.26($\pm$3.09) & 46.17($\pm$6.70) & 56.51($\pm$2.23) \\
\hline
\multirow{2}{*}{WW} 
  & \#LP & 256 & 220 & 232 & 187 & 170 & 137 & 221 & 197 & 148 & 212 \\
  & RMSE & \textbf{79.96($\pm$1.04)} & 80.67($\pm$1.43) & 80.76($\pm$1.53) & 81.13($\pm$1.60) & 80.61($\pm$1.55) & 82.32($\pm$1.33) & 80.22($\pm$1.72) & 79.98($\pm$1.74) & 80.07($\pm$1.08) & 80.05($\pm$1.25) \\
\hline
\multirow{2}{*}{CS} 
  & \#LP & 232 & 160 & 208 & 136 & 125 & 101 & 197 & 173 & 197 & 173 \\
  & RMSE & 36.99($\pm$2.83) & 36.92($\pm$1.27) & 36.67($\pm$2.07) & 38.88($\pm$1.61) & \textbf{36.01($\pm$2.48)} & 43.75($\pm$4.31) & 38.97($\pm$2.37) & 38.17($\pm$1.59) & 37.67($\pm$1.54) & 38.25($\pm$2.61) \\
\hline
\hline
Average Rank & & \textbf{2.84} & 6.84 & 3.84 & 9.34 & 6.00 & 9.17 & 4.00 & 4.17 & 4.34 & 4.50 \\
\hline
\end{tabular}
}% end \resizebox
\begin{tablenotes}
\item[(1)] RMSE values are scaled by 100.
\item[(2)] The bold values are the best results in each row.
\end{tablenotes}
\end{threeparttable}
\label{tab:bestofall_lf}
\end{table*}

\section{Conclusion and Future Work}
In this study, we introduce FAM and FAME, which leverage the strengths of additive models and SFLSs, making them ideal solutions for XAI. The proposed FAME, incorporating a novel parameterization of MFs, efficiently captures input-output relationships with fewer active rules. The presented comparative results show that both FAM and FAME are comparable to traditional MFLSs, with FAM often providing slightly higher accuracy and FAME excelling in interpretability, making it ideal for applications where explainability is crucial. Ultimately, the selection between FAM and FAME depends on the required balance between accuracy and explainability. Our findings demonstrate that these models not only perform well across various datasets but also significantly improve interpretability.

Future work will be on investigating low-dimensional space $\boldsymbol{Z} \in \mathbb{R}^{D \times 1}$  and consequent space of FAME to provide an end-to-end interpretable fuzzy family for XAI.

\section*{Acknowledgment}
The authors acknowledge using ChatGPT to refine the grammar and enhance the English language expressions.

%%%%%% Best of All L2

% \newpage
\bibliographystyle{IEEEtran}
\bibliography{IEEEabvr,cites}

% Generated by IEEEtran.bst, version: 1.14 (2015/08/26)
\begin{thebibliography}{10}
\providecommand{\url}[1]{#1}
\csname url@samestyle\endcsname
\providecommand{\newblock}{\relax}
\providecommand{\bibinfo}[2]{#2}
\providecommand{\BIBentrySTDinterwordspacing}{\spaceskip=0pt\relax}
\providecommand{\BIBentryALTinterwordstretchfactor}{4}
\providecommand{\BIBentryALTinterwordspacing}{\spaceskip=\fontdimen2\font plus
\BIBentryALTinterwordstretchfactor\fontdimen3\font minus \fontdimen4\font\relax}
\providecommand{\BIBforeignlanguage}[2]{{%
\expandafter\ifx\csname l@#1\endcsname\relax
\typeout{** WARNING: IEEEtran.bst: No hyphenation pattern has been}%
\typeout{** loaded for the language `#1'. Using the pattern for}%
\typeout{** the default language instead.}%
\else
\language=\csname l@#1\endcsname
\fi
#2}}
\providecommand{\BIBdecl}{\relax}
\BIBdecl

\bibitem{minh2022explainable}
D.~Minh, H.~X. Wang, Y.~F. Li, and T.~N. Nguyen, ``Explainable artificial intelligence: a comprehensive review,'' \emph{Artificial Intelligence Review}, pp. 1--66, 2022.

\bibitem{NEURIPS2021_251bd044}
R.~Agarwal, L.~Melnick, N.~Frosst, X.~Zhang, B.~Lengerich, R.~Caruana, and G.~E. Hinton, ``Neural additive models: Interpretable machine learning with neural nets,'' in \emph{Adv. Neural Inf. Process Syst.}, vol.~34, 2021, pp. 4699--4711.

\bibitem{9149804}
Z.~Yang, A.~Zhang, and A.~Sudjianto, ``Enhancing explainability of neural networks through architecture constraints,'' \emph{IEEE Trans. Neural Netw. Learn. Syst.}, vol.~32, no.~6, pp. 2610--2621, 2021.

\bibitem{gaminet}
------, ``Gami-net: An explainable neural network based on generalized additive models with structured interactions,'' \emph{Pattern Recognition}, vol. 120, p. 108192, 2021.

\bibitem{xnn}
J.~Vaughan, A.~Sudjianto, E.~Brahimi, J.~Chen, and V.~N. Nair, ``Explainable neural networks based on additive index models,'' \emph{arXiv preprint arXiv:1806.01933}, 2018.

\bibitem{extract}
C.~He, M.~Ma, and P.~Wang, ``Extract interpretability-accuracy balanced rules from artificial neural networks: A review,'' \emph{Neurocomputing}, vol. 387, pp. 346--358, 2020.

\bibitem{CORDON2011894}
O.~Cordón, ``A historical review of evolutionary learning methods for mamdani-type fuzzy rule-based systems: Designing interpretable genetic fuzzy systems,'' \emph{Int. J. Approx. Reason.}, vol.~52, no.~6, pp. 894--913, 2011.

\bibitem{8481251}
H.~Hagras, ``Toward human-understandable, explainable ai,'' \emph{Computer}, vol.~51, no.~9, pp. 28--36, 2018.

\bibitem{8610271}
A.~Fernandez, F.~Herrera, O.~Cordon, M.~Jose~del Jesus, and F.~Marcelloni, ``Evolutionary fuzzy systems for explainable artificial intelligence: Why, when, what for, and where to?'' \emph{IEEE Comput. Intell. Mag.}, vol.~14, no.~1, pp. 69--81, 2019.

\bibitem{9430516}
J.~M. Mendel and P.~P. Bonissone, ``Critical thinking about explainable ai (xai) for rule-based fuzzy systems,'' \emph{{IEEE} Trans. Fuzzy Syst.}, vol.~29, no.~12, pp. 3579--3593, 2021.

\bibitem{curse}
Z.~Shi, S.~Huang, L.~Wu, Q.~Zhang, X.~Zhang, Y.~Cao, Y.~Chen, and Y.~Lv, ``Tsk fuzzy system optimization for high-dimensional regression problems,'' \emph{IEEE Trans. Emerg. Top. Comput. Intell.}, 2024.

\bibitem{HTSK}
Y.~Cui, D.~Wu, and Y.~Xu, ``Curse of dimensionality for tsk fuzzy neural networks: Explanation and solutions,'' in \emph{Proc. Int. Jt. Conf. Neural Netw.}, 2021.

\bibitem{7047917}
C.~L.~P. Chen, C.-Y. Zhang, L.~Chen, and M.~Gan, ``Fuzzy restricted boltzmann machine for the enhancement of deep learning,'' \emph{{IEEE} Trans. Fuzzy Syst.}, vol.~23, no.~6, pp. 2163--2173, 2015.

\bibitem{7377034}
S.~Park, S.~J. Lee, E.~Weiss, and Y.~Motai, ``Intra- and inter-fractional variation prediction of lung tumors using fuzzy deep learning,'' \emph{IEEE J. Trans. Eng. Health. Med.}, vol.~4, pp. 1--12, 2016.

\bibitem{8015718}
S.~Rajurkar and N.~K. Verma, ``Developing deep fuzzy network with takagi sugeno fuzzy inference system,'' in \emph{IEEE International Conference on Fuzzy Systems}, 2017, pp. 1--6.

\bibitem{8491679}
R.~Chimatapu, H.~Hagras, A.~Starkey, and G.~Owusu, ``Interval type-2 fuzzy logic based stacked autoencoder deep neural network for generating explainable ai models in workforce optimization,'' in \emph{IEEE International Conference on Fuzzy Systems}, 2018.

\bibitem{10.1007/978-3-030-04070-3_1}
------, ``Explainable ai and fuzzy logic systems,'' in \emph{Theory and Practice of Natural Computing}, 2018.

\bibitem{siso_fls}
T.~Kumbasar, ``Robust stability analysis and systematic design of single-input interval type-2 fuzzy logic controllers,'' \emph{{IEEE} Trans. Fuzzy Syst.}, vol.~24, no.~3, pp. 675--694, 2016.

\bibitem{guven}
Y.~Güven, A.~Köklü, and T.~Kumbasar, ``Exploring {Z}adeh's {G}eneral {T}ype-2 {F}uzzy {L}ogic {S}ystems for {U}ncertainty {Q}uantification,'' \emph{{IEEE} Trans. Fuzzy Syst.}, 2024.

\bibitem{koklu1}
A.~Köklü, Y.~Güven, and T.~Kumbasar, ``Odyssey of interval type-2 fuzzy logic systems: Learning strategies for uncertainty quantification,'' \emph{{IEEE} Trans. Fuzzy Syst.}, pp. 1--10, 2024.

\bibitem{infus}
A.~K{\"o}kl{\"u}, Y.~G{\"u}ven, and T.~Kumbasar, ``Efficient learning of fuzzy logic systems for large-scale data using deep learning,'' in \emph{International Conference on Intelligent and Fuzzy Systems}, 2024.

\bibitem{cdr_dr}
Y.~Cui, Y.~Xu, R.~Peng, and D.~Wu, ``Layer normalization for tsk fuzzy system optimization in regression problems,'' \emph{{IEEE} Trans. Fuzzy Syst.}, vol.~31, no.~1, pp. 254--264, 2023.

\end{thebibliography}

\end{document}